\documentclass[10pt,twocolumn,letterpaper]{article}

\usepackage{cvpr}
\usepackage{times}
\usepackage{epsfig}
\usepackage{graphicx}
\usepackage{amsmath}
\usepackage{amssymb}

\usepackage{times}
\usepackage{epsfig}
\usepackage{amssymb}
\usepackage{xcolor}

\usepackage{caption}
\usepackage{subcaption}
\usepackage{fixltx2e}

\usepackage{kotex}
\usepackage{color}
\usepackage{array}

\usepackage[pagebackref=true,breaklinks=true,letterpaper=true,colorlinks,bookmarks=false]{hyperref}

\cvprfinalcopy 

\def\cvprPaperID{20} 

\newcolumntype{C}[1]{>{\centering\let\newline\\\arraybackslash\hspace{0pt}}m{#1}}

\ifcvprfinal\fi
\begin{document}

\title{Why Do Deep Neural Networks Still Not Recognize These Images?: \\ A Qualitative Analysis on Failure Cases of ImageNet Classification}

\author{Han S. Lee, Alex A. Agarwal, and Junmo Kim\\
School of Electrical Engineering, KAIST }


\maketitle


\section{Introduction}

In a recent decade, ImageNet~\cite{imagenet} has become the most notable and powerful benchmark database in computer vision and machine learning community.
With a large number of natural images and 1000+ annotated classes, ImageNet has initiated the large scale computer vision era and also has provided a chance for deep learning (DL) to achieve its golden age.
Since the DL technology has been applied, the error rates for ImageNet classification task, previously exceeding 20\%, has fallen to 10\%.
With the recent development of more advanced models such as ResNet~\cite{resnet}, DL has achieved the error rates of 3\% in 2016 ImageNet Challenge.
With this success, ImageNet's legacy is evolving eo apply DL technology into more diverse tasks, such as image captioning, visual question answering, and gaming.

As ImageNet has emerged as a representative benchmark for evaluating the performance of novel DL models, its evaluation tends to include only quantitative measures such as error rate, rather than qualitative analysis.
Recent DL models for ImageNet also has focused on how to innovate the network structure and improve the accuracy while designing a deeper and larger network.
Thus, there are few studies that analyze the failure cases of DL models in ImageNet.
In this abstract, we qualitatively analyze the failure cases of ImageNet classification results from recent DL model, and categorize these cases according to the certain image patterns.
Through this failure analysis, we believe that it can be discovered what the final challenges are in ImageNet database, which the current DL model is still vulnerable to.

\section{Experiments and Results}

To analyze the failure cases of ImageNet classification task, we propose and implement the DL model which can achieve state-of-the-art level classification error.
Our DL model is an ensemble of five DL models, including three 200-layered ResNet~\cite{resnet} and one Inception~\cite{inception} and one InceptionResNet. 
The model achieved the classification error rate of 3.29\% in the ImageNet object classification and localization task.
We then collect the failure cases from the validation set of ImageNet database to qualitatively analyze 400 images randomly selected from these failure cases.
On these failure cases, an observer monitors the entire images and suggest five categories according to the image characteristics.
The five proposed categories and their numbers are summarized in Table~\ref{tab1}.
The observer then again assess the images to classify them into the suggested categories.

\begin{table}[t!]
\centering
\caption{Number of cases for each category of failure cases in ImageNet classification task.}
\begin{tabular}{C{4cm}|C{3cm}}
\hline
Categories & \# of cases (rates) \\
\hline \hline
Similar labels (1) & 61 (15.2\%) \\
Not salient GT (2) & 87 (21.8\%) \\
Challenging images (3) & 43 (10.8\%) \\
Incorrect GT (4) & 79 (19.8\%) \\
Incorrect PC (5) & 130 (32.5\%) \\
\hline
\end{tabular}
\label{tab1}
\end{table}

\begin{figure}[t!]
	\centering
	\begin{subfigure}[b]{0.23\textwidth}
		\includegraphics[width=\textwidth]{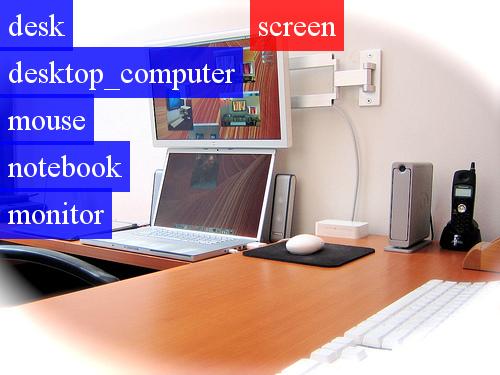}
		\label{fig2a}
	\end{subfigure}
	\begin{subfigure}[b]{0.23\textwidth}
		\includegraphics[width=\textwidth]{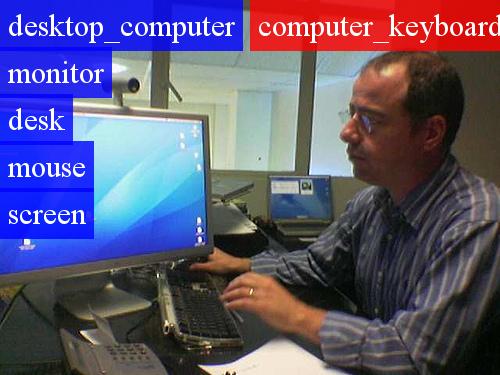}
		\label{fig2b}
	\end{subfigure}
	\begin{subfigure}[b]{0.23\textwidth}
		\includegraphics[width=\textwidth]{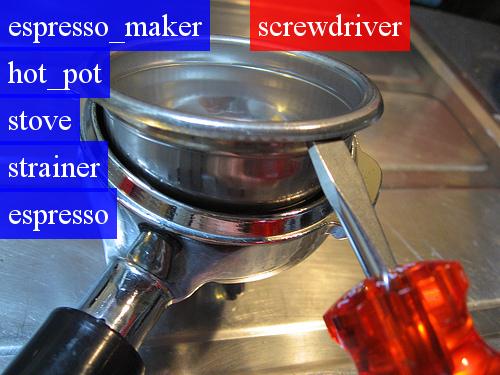}
		\label{fig2a}
	\end{subfigure}
	\begin{subfigure}[b]{0.23\textwidth}
		\includegraphics[width=\textwidth]{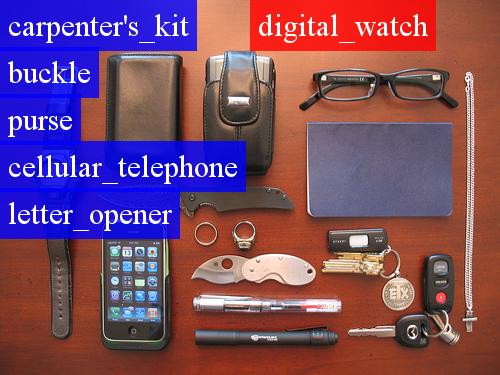}
		\label{fig2b}
	\end{subfigure}
	\begin{subfigure}[b]{0.23\textwidth}
		\includegraphics[width=\textwidth]{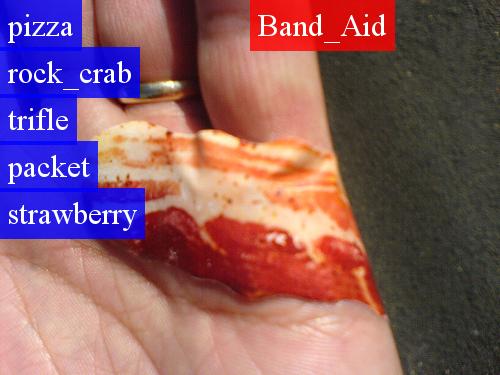}
		\label{fig2a}
	\end{subfigure}
	\begin{subfigure}[b]{0.23\textwidth}
		\includegraphics[width=\textwidth]{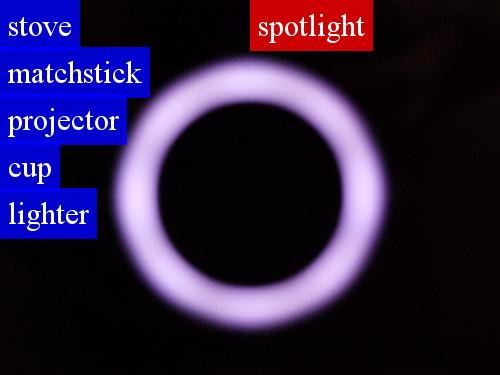}
		\label{fig2c}
	\end{subfigure}
	\begin{subfigure}[b]{0.23\textwidth}
		\includegraphics[width=\textwidth]{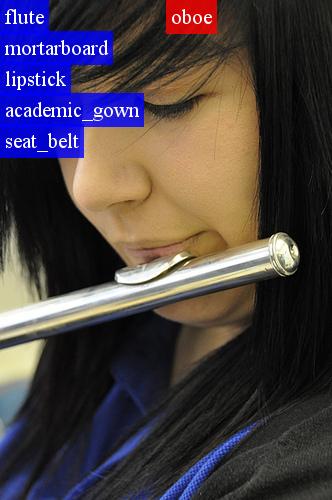}
		\label{fig2a}
	\end{subfigure}
	\begin{subfigure}[b]{0.23\textwidth}
		\includegraphics[width=\textwidth]{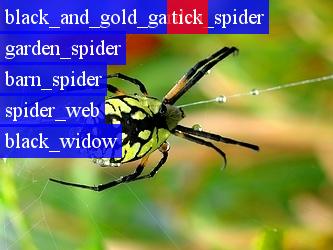}
		\label{fig2c}
	\end{subfigure}
	\begin{subfigure}[b]{0.23\textwidth}
		\includegraphics[width=\textwidth]{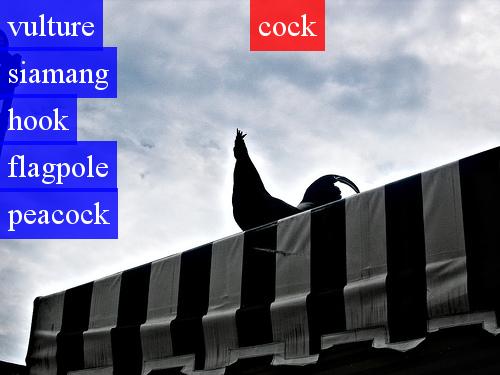}
		\label{fig2b}
	\end{subfigure}
	\begin{subfigure}[b]{0.23\textwidth}
		\includegraphics[width=\textwidth]{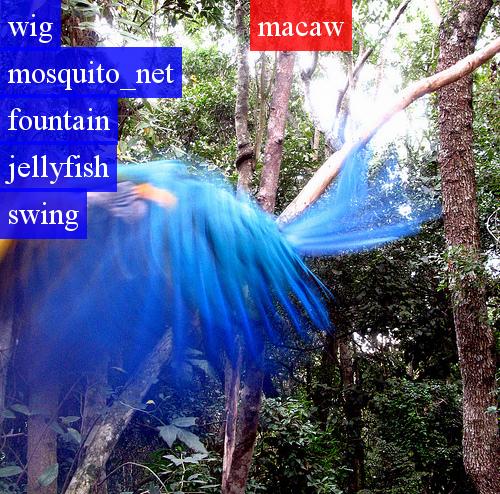}
		\label{fig2c}
	\end{subfigure}
	\caption{Example images and their GT (red) and PC (blue) of ImageNet failure cases. Each row includes the cases of one category from; (1) Similar labels, (2) not salient GT, (3) challenging images, (4) incorrect GT, and (5) incorrect PC. (from the top row)}
	\label{fig1}
\end{figure}

Fig.~\ref{fig1} demonstrates some failure case examples with regard to the suggested categories.
First, \textit{Similar Labels} are the cases where the ground truths (GT) and the predicted classes (PC) refer to almost similar contents.
In this category, PC are not wrong and have the same or similar meaning to the GT, so that these are more of an evaluation error than a failure.
For instance, numerous similar class words such as ``monitor,'' ''desktop computer,'' ``computer keyboard,'' and ''notebook'' which are all related to the computers on the desk, appear duplicated in the images in the first row of Fig.~\ref{fig1}.
In order to handle with these failure cases, the evaluation policy should be revised to correctly evaluate the classification of similar semantic classes, from the database perspective.
Moreover, from the DL model perspective, it is necessary to improve the model to consider the inter-class similarity, so that when some classes obtain high prediction weights, their similar classes also do.

Second, \textit{Not Salient GT} are the cases where it is hard to say that PC is incorrect because GT is not salient in the image.
As shown in the second row of Fig.~\ref{fig1}, cases were found in which GT exists in the images but not are salient, and PC also exists and sometimes even are salient.
These cases are considered to be supplemented by revising the GT of the database or improving the evaluation policy to include multiple GT.
Third, \textit{Challenging Images} refer to the images that are particularly difficult to predict the objects even by humans.
The bacon-like bandage from the third row in Fig.~\ref{fig1} is quite difficult to recognize even by human observer without the contextual knowledge.
Alternatives are not considered because we believe it can be solved by increasing the learning complexity of DL, which is beyond the scope of this abstract.

Fourth, \textit{Incorrect GT} refer to the cases where GT is simply wrong.
Surprisingly, almost 20\% of the failure cases were caused by incorrect GT, such as annotating flute as oboe or spider as tick, as in fourth row of Fig.~\ref{fig1}.
One interesting fact is that the majority of animal-related images in the failure cases fall into this category, which can lead to the conclusion that the performance of DL overwhelms that of humans in the animal classification task.
This incorrect GT issue should be complemented by the strict revision of GT in the database.
Finally, \textit{Incorrect PC} are the cases where PC is wrong.
The 1st, 2nd, and 4th categories are the cases in which the PC actually performed the classification correctly.
In this sense, the 3rd and 5th, especially 5th category, can be seen as a clear failure of the DL in ImageNet classification.
However, As seen in the 5th row of Fig.~\ref{fig1}, the images in the most of the category is out of its regular range of imaging distribution, with the distortion of illumination and motion blur.
In order to classify these correctly, DL can be improved by training distorted images in which the regular training images are modified by these irregular illumincation and motion blur.

In this abstract, we collected the failure cases of the recent DL model in ImageNet database, categorized them into case-by-case via qualitative analysis, and discussed their problems and alternatives.
It is expected that our work will allow DL researchers to discover the remaining challenges when applying DL to various large scale image classification databases, not only ImageNet, and also allow database coordinators to know what points to keep in mind when building new databases.
Future works include the further analysis on these categorized failure cases and improving DL by considering the above challenges such as inter-class similarity and training distorted images.

{\small
\bibliographystyle{ieee}
\bibliography{egbib}
}

\end{document}